\documentclass[conference]{IEEEtran}
\IEEEoverridecommandlockouts
\usepackage{cite}
\usepackage{amsmath,amssymb,amsfonts}
\usepackage{algorithmic}
\usepackage{graphicx}
\usepackage{textcomp}
\usepackage[table]{xcolor}
\usepackage{booktabs}
\def\BibTeX{{\rm B\kern-.05em{\sc i\kern-.025em b}\kern-.08em
    T\kern-.1667em\lower.7ex\hbox{E}\kern-.125emX}}
\begin{document}

\title{Scaling Latent Motor Adaptation to Coordinated Bimanual Manipulation}

\author{
\IEEEauthorblockN{
Seung-Bin Joo\IEEEauthorrefmark{1},
Alessandro Albini\IEEEauthorrefmark{2},
Luca Wiehe\IEEEauthorrefmark{1}\IEEEauthorrefmark{3},
Perla Maiolino\IEEEauthorrefmark{2},
Daniel Olmeda Reino\IEEEauthorrefmark{4},
João F. Henriques\IEEEauthorrefmark{1}
}
\IEEEauthorblockA{
\IEEEauthorrefmark{1}Visual Geometry Group,
University of Oxford, Oxford, United Kingdom\\
\IEEEauthorrefmark{2}Soft Robotics Lab, Oxford Robotics Institute,
University of Oxford, Oxford, United Kingdom\\
\IEEEauthorrefmark{3}Technical University of Munich, Munich, Germany\\
\IEEEauthorrefmark{4}AI Core Research,
Toyota Motor Europe, Brussels, Belgium\\
Email: \texttt{joo@robots.ox.ac.uk}
}
}

\maketitle

\begin{abstract}
Robust manipulation under uncertain object dynamics and geometry remains a core challenge in real-world robotics. Policies trained under fixed assumptions often degrade when object mass, friction, scale, or shape vary. Embedding-based adaptation methods such as Rapid Motor Adaptation (RMA) address this by conditioning actions on latent representations of environment dynamics inferred from interaction history. We investigate whether this paradigm scales from single arm settings to coordinated bimanual manipulation, where higher dimensional control, coupled contacts, and shared object forces introduce additional complexity. We evaluate bimanual RMA on three contact-rich ManiSkill tasks with randomized YCB object properties. Bimanual RMA, using observation-action history and dual wrist depth images, retains privileged policy performance on bimanual picking and lifting, while an otherwise matched stationary PPO policy without latent conditioning achieves low success; all methods remain limited on long-horizon stacking. We further demonstrate direct sim-to-real transfer on a dual Franka lifting task without real-world finetuning. Together, these results suggest that latent conditioning provides useful structure for scaling robust control to coordinated bimanual systems, while longer horizon tasks require additional temporal structure.

\end{abstract}

\section{Introduction}

For robotic manipulation to operate reliably outside controlled settings, policies must cope with variation in object geometry and dynamics \cite{mu2021maniskillgeneralizablemanipulationskill}. Object mass, friction, scale, and shape are often unknown or only approximately modeled at deployment, causing policies trained for a fixed configuration to degrade. Embedding-based system identification offers one approach to this problem. Rapid Motor Adaptation (RMA) \cite{kumar2021rmarapidmotoradaptation} conditions a policy on a learned latent representation of the environment inferred from interaction history. By separating privileged policy training from deployment-time embedding prediction, RMA allows behavior to specialize across dynamics regimes without explicitly estimating physical parameters. This paradigm has been demonstrated in locomotion, in-hand rotation, and single-arm manipulation \cite{kumar2021rmarapidmotoradaptation, kumar2022adaptingrapidmotoradaptation, liang2024rmaRobotManipulatorArms,qi2023rmaInHandRotation}.

However, it remains unclear whether latent adaptation scales to coordinated bimanual manipulation. Dual-arm systems introduce higher dimensional control, coupled multi-contact interactions, shared object forces, and sensitivity to asymmetric contacts \cite{grotz2024peract2benchmarkinglearningrobotic, karim2024daviladaptivedualarmmanipulation, grannen2023stabilizeactlearningcoordinate}. In tasks such as lifting, handover, and stacking, the action of one arm changes the object state and contact forces experienced by the other. Object uncertainty thus interacts directly with coordination, raising the question of whether scaling an existing adaptation framework is sufficient or whether bimanual control requires additional structure.

\begin{figure}[t!]
    \centering
    \includegraphics[width=\columnwidth]{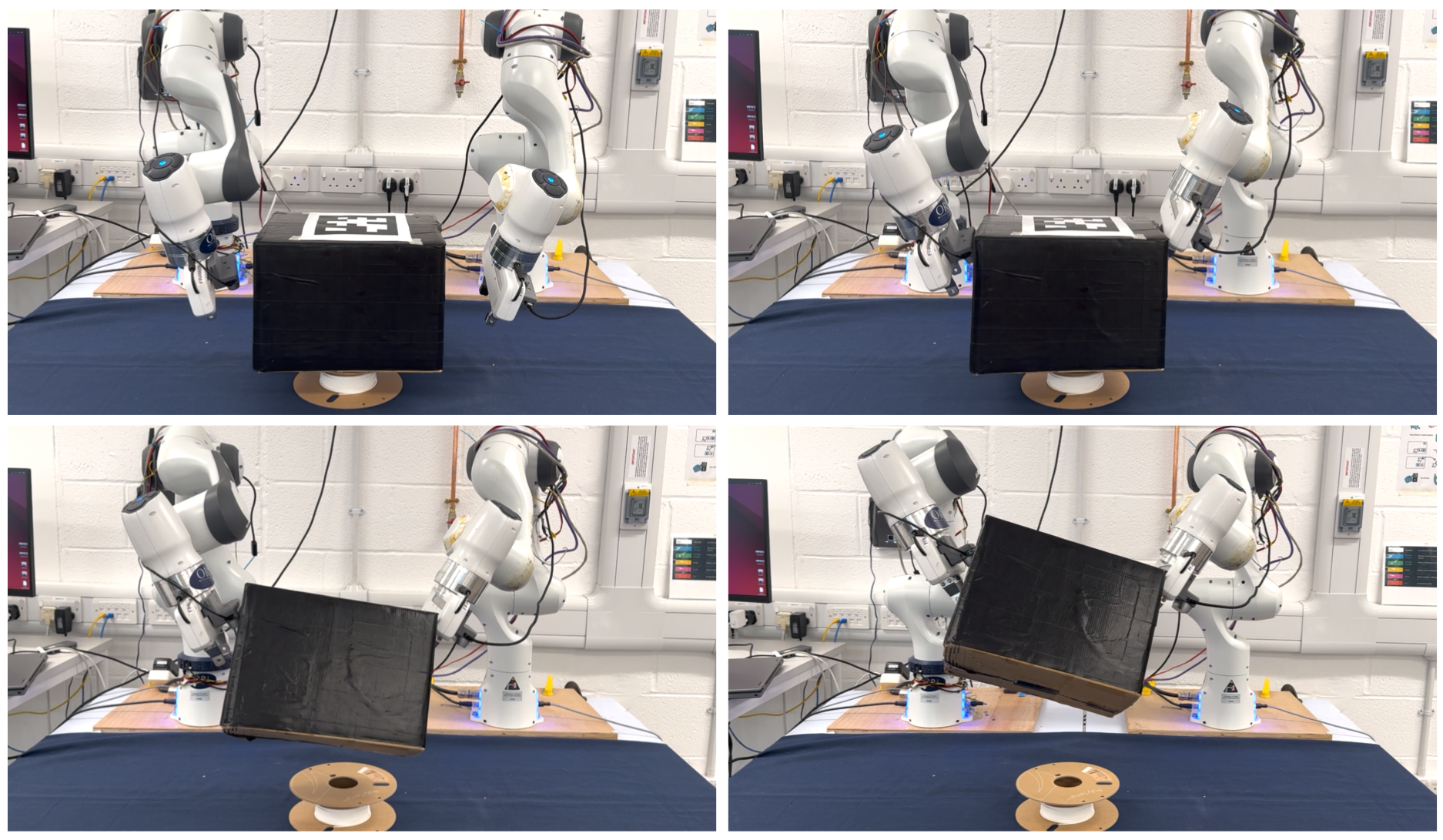}
    \caption{We train bimanual visuo-motor policies in simulation that use intra-episode latent parameter inference and can be deployed in the real-world. Our system supports both manipulation of a variety of household objects and more complex lifting tasks with real robots as shown in the above sequence.}
    \label{exp_irl}
\end{figure}

We empirically investigate this question by applying RMA to centralized bimanual control under heterogeneous object geometry and dynamics. We retain the original RMA formulation: a single policy produces 14-DoF actions conditioned on a shared environment embedding, while a deployment adapter predicts this embedding from bimanual observation-action history and dual wrist depth images. Our investigation is therefore empirical rather than algorithmic: we study the applicability and limitations of latent environment conditioning in contact-rich bimanual manipulation. Our main contributions are as follows. \textbf{(1)} We empirically investigate whether latent environment conditioning can scale from single-arm settings to centralized 14-DoF bimanual control under heterogeneous object geometry and dynamics. We evaluate three custom ManiSkill tasks involving YCB objects with continuous variation in scale, density, and friction. \textbf{(2)} The bimanual RMA policy retains privileged policy performance on picking and lifting and outperforms an otherwise matched PPO policy without latent conditioning, while long-horizon stacking remains challenging. \textbf{(3)} We demonstrate direct sim-to-real deployment of embedding-conditioned bimanual control on two Franka arms, achieving 70\% success over 20 lifting trials without real-world finetuning.

Together, these results suggest that latent environment conditioning provides useful structure for scaling robust control to coordinated bimanual systems, while longer-horizon tasks require additional temporal and coordination structure.

\section{Related Work} \label{sec:related}
\paragraph{Learning-based manipulation} Learning-based manipulation broadly follows scaling and interaction-driven approaches. Imitation learning (IL), including vision-language-action models, achieves strong single and dual-arm performance by scaling transformer policies and demonstrations \cite{black2024pi0}. However, these methods typically cannot improve through interaction and require high quality data, particularly costly for bimanual systems needing coordinated teleoperation across diverse objects \cite{shaw2024bimanualdexteritycomplextasks,zhao2023learningfinegrainedbimanualmanipulation}. Reinforcement learning (RL) instead learns through interaction \cite{kalashnikov2018qtoptscalabledeepreinforcement,gualtieri2018pickplacegeometricobject}, often using domain randomization for robustness and sim-to-real transfer \cite{tobin2017domainrandomizationtransferringdeep,Peng_2018,handa2024dextremetransferagileinhand,openai2019solvingrubikscuberobot}. Online system identification conditions policies on dynamics inferred from interaction histories \cite{yu2017preparingunknownlearninguniversal}, while RMA replaces explicit parameter estimation with a learned environment embedding and has been applied to locomotion, in-hand rotation, and single arm manipulation \cite{kumar2021rmarapidmotoradaptation,kumar2022adaptingrapidmotoradaptation,qi2023rmaInHandRotation,liang2024rmaRobotManipulatorArms}.

\paragraph{Bimanual manipulation} Bimanual manipulation adds higher dimensional control, inter-arm coupling, and shared contact constraints. Classical methods used optimization-based coordination \cite{SMITH20121340}, while recent learning-based work spans complementary scaling and structural perspectives. Scaling-oriented systems such as \textit{ALOHA Unleashed} combine large bimanual demonstration datasets with diffusion policies \cite{zhao2024alohaunleashedsimplerecipe}; related IL methods use voxel-based representations and transformer architectures \cite{liu2024voxactb,grotz2024peract2benchmarkinglearningrobotic}. RL approaches combine interaction, sim-to-real techniques, and algorithms such as PPO \cite{lin2025simtorealreinforcementlearningvisionbased,chen2022humanlevelbimanualdexterousmanipulation}, while structural approaches may assign one arm a stabilizing role and the other the primary action \cite{grannen2023stabilizeactlearningcoordinate}. Our work connects these views by combining dynamics randomization with a centralized policy conditioned on a shared inferred latent, testing whether this lightweight structure extends RMA to bimanual manipulation.

\section{Bimanual Latent Adaptation} \label{sec:method}

\begin{figure}[t]
    \centering
    \includegraphics[width=0.9\columnwidth]{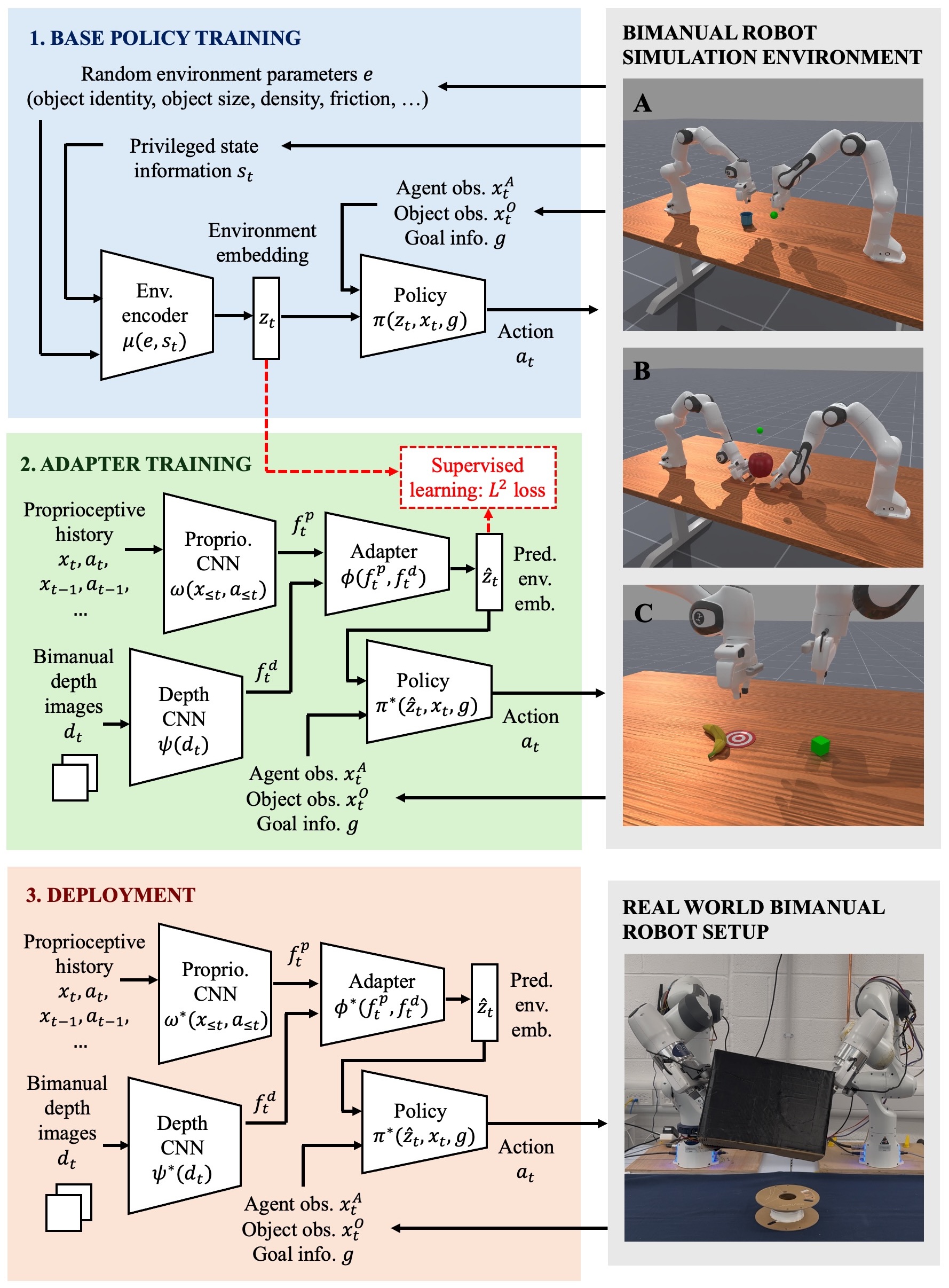}
    \caption{Overview of RMA pipeline for bimanual manipulators. During base policy training, the environment encoder $\mu(e,s_t^{\mathrm{priv}})$ embeds the privileged information from simulation into vector $z_t$, and the policy $\pi (z_t, x_t, g)$ is conditioned on this environment embedding $z_t$. During adapter training, the base policy $\pi^{*} (z_t, x_t, g)$ is frozen, and the adapter $\phi (f_t^p,f_t^d)$, proprioception CNN $\omega(x_{\leq t}, a_{\leq t})$ and depth CNN $\psi(d_t)$ are trained jointly to predict the environment embedding $\hat{z}_t$ via MSE regression to the ground truth $z_t$. During deployment, we generate actions using the optimized adapters and base policy. We also show \textit{Bimanual Pick YCB} (A), \textit{Bimanual Lift YCB} (B), and \textit{Bimanual Stack YCB} (C) task environments.} \label{RMA diagram}
\end{figure}

We apply RMA \cite{kumar2021rmarapidmotoradaptation} to bimanual manipulation without modifying its core formulation. As shown in Fig.~\ref{RMA diagram}, during privileged training, an encoder maps object properties and privileged state to a shared latent $z_t$ that conditions a dual arm policy. Domain randomization exposes the policy to diverse object configurations, while $z_t$ conditions its bimanual behavior on object geometry and dynamics. The policy and encoder are then frozen, and an adapter learns to reconstruct $z_t$ from bimanual proprioceptive history and wrist depth observations. At deployment, the adapter replaces the privileged encoder, enabling the policy to select behavior appropriate to inferred object properties without simulator-only information.

\subsection{Privileged Base Policy Training}

We formulate each task as an MDP with simulator state $s_t$, deployable observation $x_t=o(s_t)$, task goal $g$, and environment parameters $e$. An encoder $\mu$ maps privileged information to an environment embedding $z_t$ that conditions the policy $\pi$. They are jointly optimized as
\begin{equation} \label{base_policy_eq}
        \pi^*, \mu^* = \underset{\pi, \mu}{\operatorname{argmax}}\; \mathop{\mathbb{E}}\limits_{\substack{e \sim \mathcal{E} \\ g \sim \mathcal{G}}} \left[ \mathop{\mathbb{E}}\limits_{s_{t+1} \sim P_e(\cdot | s_t, a_t)} \left[ \sum_{t'=0}^{T-1} \gamma^{t'} r(s_{t'}, g) \right] \right]
\end{equation}
\begin{align*}
    & \text{with } s_{t+1}\sim P_e(\cdot\mid s_t,a_t), \quad x_t = o(s_t), \\
    & a_t \sim \pi(x_t, z_t, g), \quad z_t=\mu(e,s_t^{\mathrm{priv}}).
\end{align*}
Here, $\mathcal{G}$ is the task's goal distribution, with $g$ specifying desired object pose. The policy observation $x_t$ contains joint positions and velocities, TCP poses, object position, and bimanual TCP-to-object displacements. These variables are available at deployment, with object position supplied by the simulator during training and an external tracker on hardware. The encoder additionally receives privileged state $s_t^{\mathrm{priv}}$, comprising object orientation and gripper-object contact impulses.

Environment $e\sim\mathcal{E}$ is constructed by selecting an object instance and category and randomizing its scale, density, and friction. Instead of explicitly describing object geometry, we map instance and category indices to learned embeddings \cite{liang2024rmaRobotManipulatorArms}:
\begin{equation}
e=\operatorname{concat}
\left(e_{\mathrm{obj}},e_{\mathrm{category}},e_{\mathrm{physical}}\right),
\label{env_param_eq}
\end{equation}
where $e_{\mathrm{physical}}$ contains scale, density, and friction. The instance and category embeddings implicitly represent object geometry, while the physical parameters are provided directly. The encoder compresses $e$ and $s_t^{\mathrm{priv}}$ into the latent $z_t$, rather than estimating interpretable physical parameters. The policy and encoder are MLPs trained jointly using PPO \cite{schulman2017proximalpolicyoptimizationalgorithms}.

We use one centralized policy to control both 7-DoF arms. Bimanual observations, actions, and privileged contact information are concatenated, and both arms are conditioned on the same object latent. Unlike independent-agent or leader--follower formulations \cite{chen2022humanlevelbimanualdexterousmanipulation,jiang2025rethinkingbimanualroboticmanipulation,grotz2024peract2benchmarkinglearningrobotic}, our formulation assigns no predefined arm roles while allowing asymmetric, coordinated motions through a joint 14-DoF action.

\subsection{Adapter Training}
\label{sec}

Because privileged environment information is unavailable outside simulation, we train an adapter to predict the embedding $z_t$ required by the frozen base policy. A 1D CNN $\omega$ processes the concatenated bimanual observation-action history $(x_{\leq t},a_{\leq t})$, encoding how the robot states evolve under applied actions and thereby providing information about latent physical properties such as object density and friction. Since proprioception alone provides limited geometry information, two $32\times32$ wrist-depth images $d_t^1$ and $d_t^2$ are processed by a shared CNN $\psi$ to obtain coarse multi-view geometry features $f_t^d$ \cite{liang2024rmaRobotManipulatorArms}. The bimanual images share encoder weights and are processed together along the batch dimension, maintaining a compact architecture. The adapter $\phi$ combines $f_t^p$ and $f_t^d$ to predict $\hat z_t$. We jointly train $\omega$, $\psi$, and $\phi$ on simulated trajectories by minimizing the $L^2$ error between $\hat z_t$ and the privileged target $z_t=\mu^*(e,s_t^{\mathrm{priv}})$.

\subsection{Task-Specific Multi-Stage Bimanual Reward} \label{lift_reward}

Sparse exploration was unreliable for contact-rich bimanual lifting, so we use a task-specific gated reward, which encodes under-object contact $\rightarrow$ lift $\rightarrow$ stabilize $\rightarrow$ success. Let $p_i^z$ and $p_o^z$ denote the heights of arm $i$'s TCP and the object, respectively, and let $s$ estimate the object size. The binary term $r_{\mathrm{uc}}=1$ when both TCPs satisfy $p_i^z<p_o^z-s/2$ and every finger $f_i$ and object $o$ contact impulse satisfies $\lVert I(f_i,o)\rVert>\epsilon$; otherwise, $r_{\mathrm{uc}}=0$. Given object and goal positions $p_o$ and $p_g$, lifting progress is $r_{\mathrm{lift}}=1-\tanh(5\lVert p_g-p_o\rVert)$. Grasp stability is $r_{\mathrm{static}}=\frac{1}{2}\sum_{i=1}^{2}[1-\tanh(5\lVert\dot q_i\rVert)]$, where $\dot q_i$ is the joint-velocity vector of arm $i$. Finally, $z_{\mathrm{above}}=1$ when the object exceeds the target height while retaining all finger contacts, and is zero otherwise. The overall reward is
\begin{equation}
r=2r_{\mathrm{uc}}+3r_{\mathrm{uc}}r_{\mathrm{lift}}
+z_{\mathrm{above}}r_{\mathrm{static}}+5r_{\mathrm{success}},
\label{eq:lift_reward}
\end{equation}
where $r_{\mathrm{success}}$ is the binary task success reward. The multiplicative gates activate lifting only after under-object contact, and stabilization only after reaching the target height. This reward is lifting-specific rather than a general reward design contribution.

\section{Simulation and Training Setup}

\paragraph{Bimanual tasks with diverse objects}
Using ManiSkill \cite{tao2024maniskill3gpuparallelizedrobotics}, we construct 3 bimanual tasks with objects sampled from the 78 object YCB dataset \cite{7251504}. In \textit{Bimanual Pick YCB} (Fig.~\ref{RMA diagram}A), the object is outside the right arm's reach and the goal outside the left arm's reach, requiring coordinated transfer; object $z$-rotation and $xy$-position and goal $xyz$-position are randomized. In \textit{Bimanual Lift YCB} (Fig.~\ref{RMA diagram}B), both arms lift 1 of 10 round YCB objects, scaled beyond the capacity of a single gripper, above a target height; object $z$-rotation is randomized and we use the reward in Sec.~\ref{lift_reward}. In \textit{Bimanual Stack YCB} (Fig.~\ref{RMA diagram}C), the right arm places a cube at the target while the left moves a sampled YCB object to make space and subsequently stacks it on the cube. Object rotations and positions and the target location are randomized. Pick and Stack adapt dense rewards from ManiSkill's corresponding two robot tasks to diverse YCB objects.

\paragraph{Simulation setup}
Each task uses two 7-DoF Franka Emika Panda arms with torque sensing and two-finger grippers. A PD controller receives 14-dimensional actions specifying desired joint-position changes. Object scale, density, and friction are randomized by factors of $0.7$-$1.2$, $0.5$-$5.0$, and $0.5$-$1.1$ relative to their defaults, respectively, together with external disturbances up to $2.0$N. Randomization ranges are linearly increased from their default values during training. We train in 256 parallel heterogeneous environments and evaluate in 512 environments.

\paragraph{Training and architecture}
Base policies are trained for 150M simulation steps on Pick and 70M on Lift and Stack using PPO with clip coefficient $0.2$, learning rate $3\times10^{-4}$, discount $\gamma=0.8$, and Adam \cite{kingma2017adammethodstochasticoptimization}. The adapter is trained for 2M steps using the embedding MSE objective. Actor and critic networks are four-layer, width-256 $\tanh$ MLPs, while the privileged encoder is a three-layer, width-128 ELU MLP with LayerNorm. A two-layer ReLU adapter fuses a 50-step proprioceptive history, encoded by a per-step MLP and temporal convolutions with kernels $(9,7,5,3)$, with wrist depth encoded by three convolutional layers with $(32,64,128)$ channels and fully connected layers of sizes $(256,64)$.

\section{Simulation Results and Discussion}

\begin{figure}[t]
    \centering
    \includegraphics[width=0.8\columnwidth]{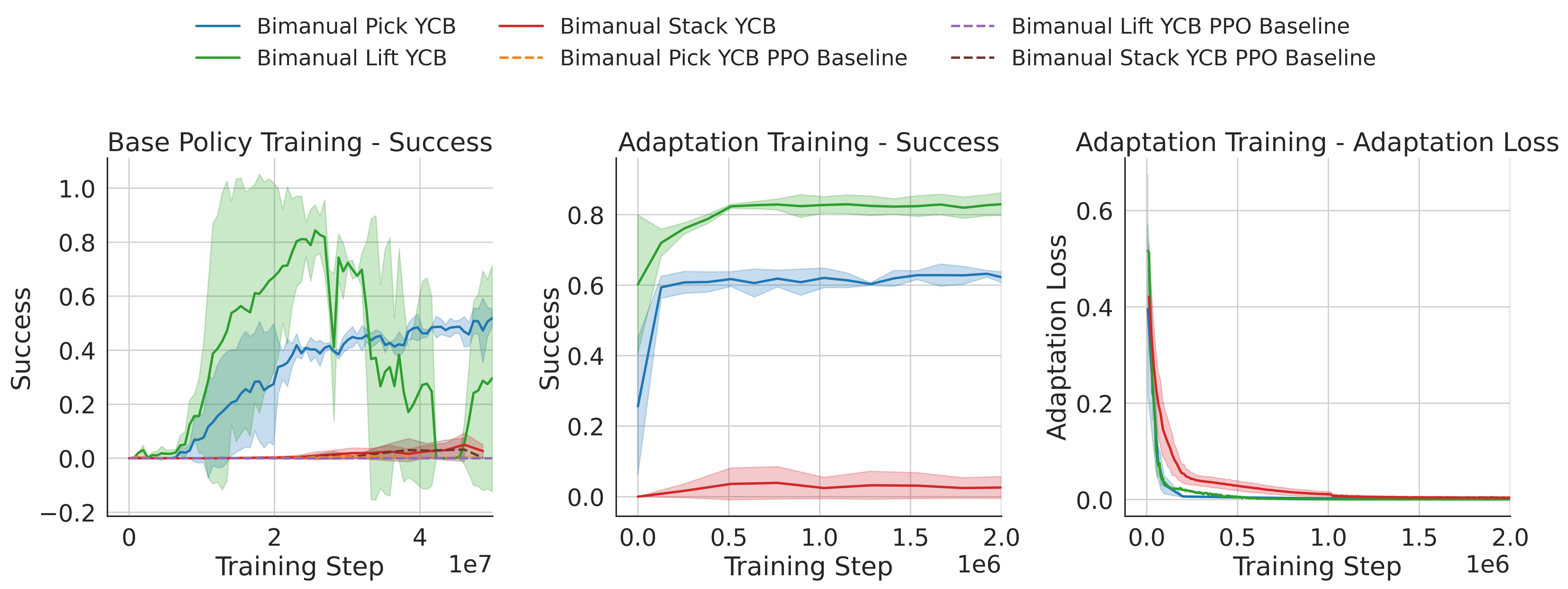}
    \caption{Learning curves for all three tasks. Left: base policy success over 50M PPO steps. Middle: success of the frozen policy conditioned on the predicted latent during adapter training. Right: adapter loss, measured as MSE between privileged and predicted embeddings. Adapter training uses the best base-policy checkpoint after domain randomization reaches its maximum. Curves show the mean and std over four seeds.}
    \label{fig:training_plots}
\end{figure}

Figure~\ref{fig:training_plots} shows base policy and adapter learning for all three tasks, where success is the fraction of 512 evaluation environments satisfying the task condition at least once per episode. For adapter training, we select the highest success privileged checkpoint after domain randomization reaches its maximum. Adapter loss decreases rapidly, and within 2M steps the deployed policy recovers performance close to its privileged teacher. As shown in Table~\ref{tab:bimanual_robot_results_resized}, RMA achieves 65.74\% and 85.13\% success on Pick and Lift, closely matching the privileged oracle at 66.17\% and 87.60\%, respectively. The otherwise matched PPO baseline (with identical domain randomization and hyperparameters), differing only by the absence of latent conditioning, achieves 1.16\% and 0\%. Stack remains difficult for all methods: the privileged policy, RMA, and PPO reach only 5.93\%, 3.81\%, and 2.73\%, respectively.

\begin{figure}[t]
    \centering
    \includegraphics[width=0.7\linewidth]{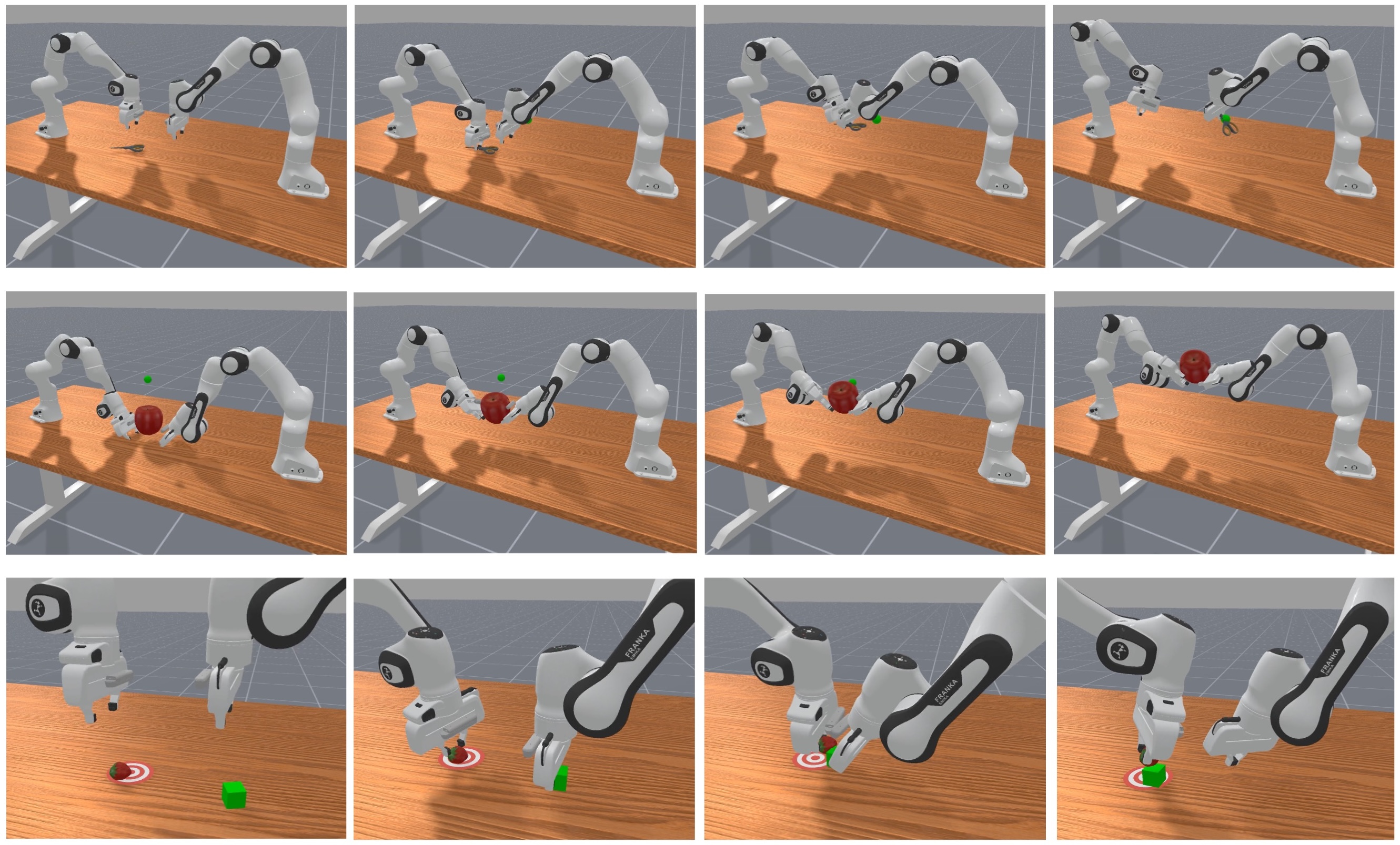}
    \caption{Example successful episodes: \emph{Bimanual Pick YCB} (top row), \emph{Bimanual Lift YCB} (middle row), \emph{Bimanual Stack YCB} (bottom row).}
    \label{fig:episodes}
\end{figure}

Under the matched training recipe, the unconditioned PPO baseline fails on Pick and Lift, whereas the latent-conditioned policy closely matches the privileged policy. This supports the value of environment-dependent conditioning in this setting, although stronger baselines are needed to separate representation and optimization effects. Qualitatively, RMA coordinates manipulation of complex objects such as scissors (Figure~\ref{fig:episodes}, row 1) and learns asymmetric motions, including opposing end-effector rotations that stabilize round objects during lifting (Figure~\ref{fig:episodes}, row 2). Wrist depth provides geometric cues during approach; after contact, close-range depth views are often degraded, making interaction history particularly important. Conversely, Stack suggests that environment latent conditioning alone is insufficient for long-horizon sequencing and precise placement. Thus, the shared latent provides useful coordination structure for Pick and Lift, but more explicit temporal or hierarchical structure may be required for longer-horizon bimanual tasks.


\begin{table}[t]
  \centering
    \caption{Simulation success rates (\%, mean$\pm$std over four seeds)
    }
  \label{tab:bimanual_robot_results_resized}
  \resizebox{\linewidth}{!}{%
    \begin{tabular}{lccc}
      \toprule
      & Bimanual Pick YCB & Bimanual Lift YCB & Bimanual Stack YCB \\
      \midrule
        \textcolor{black!50}{\textit{Privileged policy (oracle)}}
        & \textcolor{black!50}{\textit{66.17 $\pm$ 0.74}}
        & \textcolor{black!50}{\textit{87.60 $\pm$ 5.40}}
        & \textcolor{black!50}{\textit{5.93 $\pm$ 3.97}} \\
      PPO w/ Domain Randomization
      & 1.16 $\pm$ 0.98
      & 0.00 $\pm$ 0.00
      & 2.73 $\pm$ 3.04 \\
      \textbf{Bimanual RMA (ours)}
      & \textbf{65.74 $\pm$ 0.88}
      & \textbf{85.13 $\pm$ 4.58}
      & \textbf{3.81 $\pm$ 4.42} \\
      \bottomrule
    \end{tabular}%
  }
\end{table}

\section{Real-World Experiments}
\label{sec:real-world}

We evaluate direct sim-to-real transfer on two Franka Emika Panda arms lifting a $0.232\times0.320\times0.230 \mathrm{m}$ box to $0.35 \mathrm{m}$, as shown in Fig.~\ref{exp_irl}. Each wrist carries an Intel RealSense D435i depth camera, while an external camera and AprilTag provide the box position used by the policy; wrist depth is used only for latent inference. The policy is trained entirely in simulation using the measured relative robot poses and the domain randomization described above, without real-world finetuning. The adapter processes a 50-step proprioceptive history and both depth views at 10Hz, while the policy produces 14-dimensional joint-position changes at 50Hz using the latest latent estimate. Commands are executed by a compliant impedance controller at 1kHz, with joint stiffness $\mathbf{K}=[120,120,120,120,50,30,10]$ and damping $\mathbf{D}=[10,10,10,10,6,5,3]$ for each arm.

We conduct 20 trials with varied initial arm configurations and box poses. Success requires lifting the box above $0.35\mathrm{m}$ and reducing both arms' joint-velocity norms below $0.2$ at completion. The policy succeeds in 14/20 trials (70\%). Of the 6 failures, 2 result from insufficient grip force and object slippage before reaching the target, while 4 reach the target height but fail the stability criterion because of residual oscillation.

\section{Limitations and Scope}
\label{sec:limitations}

Our evaluation covers randomized scale, density, and friction over a predefined object set, rather than open-world generalization to unseen semantic categories. Object pose is supplied by simulation or external tracking, and the lifting experiments use approximately round objects and a task-specific reward. Hardware validation is limited to one box lifting task, where slippage and residual oscillation expose limitations in force modulation and terminal stabilization. These failures motivate force-aware conditioning and further stabilization and safety analysis. Finally, this is an empirical application of RMA rather than a new adaptation algorithm; broader geometries, tasks, baselines, and ablations are required to establish full generality.

\section{Conclusion}

We investigated whether latent environment conditioning scales to centralized 14-DoF bimanual manipulation under heterogeneous object dynamics. Across 3 randomized ManiSkill tasks, the deployed adapter closely retained privileged policy performance on Pick and Lift and substantially outperformed stationary PPO, although all methods remained limited on longer-horizon Stack. We further achieved 70\% success across 20 dual Franka lifting trials without real-world finetuning. These results suggest that domain randomized training combined with a shared environment latent provides useful lightweight structure for scaling adaptation to coordinated bimanual control. However, longer-horizon and force-sensitive tasks likely require additional temporal, hierarchical, or contact-aware coordination structure.

\noindent{\bf Acknowledgements.} The authors thank the support of Toyota Motor Europe (TME) and EPSRC (VisualAI, EP/T028572/1).

\bibliographystyle{IEEEtran}
\bibliography{bibliography}

\end{document}